%% file: main.tex
\documentclass[letterpaper]{article} 
\usepackage{aaai24}  
\usepackage{times}  
\usepackage{helvet}  
\usepackage{courier}  
\usepackage[hyphens]{url}  
\usepackage{graphicx} 
\usepackage{natbib}  
\usepackage{caption} 
\usepackage{algorithm2e}
\usepackage{booktabs}

\usepackage{newfloat}
\usepackage{listings}
\DeclareCaptionStyle{ruled}{labelfont=normalfont,labelsep=colon,strut=off} 
\usepackage{todonotes}

\title{Reinforcement Learning and Data-Generation for Syntax-Guided Synthesis}
\author{
    Julian Parsert\textsuperscript{\rm 1 2},
    Elizabeth Polgreen\textsuperscript{\rm 2},
}
\affiliations{
    \textsuperscript{\rm 1}University of Oxford\\
    \textsuperscript{\rm 2}University of Edinburgh\\

    julian.parsert@gmail.com, elizabeth.polgreen@ed.ac.uk
}

\input{macros}

\begin{document}

\maketitle

\begin{abstract}
Program synthesis is the task of automatically generating code based on a specification. In Syntax-Guided Synthesis (SyGuS) this specification is a combination of a syntactic template and a logical formula, and the result 
is guaranteed to satisfy both. 
We present a reinforcement-learning guided algorithm for SyGuS which uses Monte-Carlo Tree Search (MCTS) to search the space of candidate solutions. Our algorithm learns policy and value functions which, combined with the upper confidence bound for trees, allow it to balance exploration and exploitation. 
A common challenge in applying machine learning approaches to syntax-guided synthesis is the scarcity of training data. To address this, we present a method for automatically generating training data for SyGuS based on \emph{anti-unification} of existing first-order satisfiability problems, which we use to train our MCTS policy. We implement and evaluate this setup and demonstrate that learned policy and value improve the synthesis performance over a baseline by over $26$ percentage points in the training and testing sets. Our tool outperforms state-of-the-art tool cvc5 on the training set and performs comparably in terms of the total number of problems solved on the testing set (solving $23\%$ of the benchmarks on which cvc5 fails).
We make our data set publicly available, to enable further application of machine learning methods to the SyGuS problem.
\end{abstract}


\section{Introduction}\label{sec:intro}
Syntax-Guided Synthesis~\cite{sygus} allows a user to synthesize code that is guaranteed to satisfy a logical and a syntactic specification, as shown in \prettyref{ex:sygus-ex}. The logical specification is given as a quantified formula: in the example, we must find a program such that $\forall x,y. f(x,y)\geq x \wedge f(x,y)\geq y \wedge f(x,y)=x \vee x(f,y)= y$. The syntactic specification is given as a context-free grammar. 
Applications of SyGuS include planning~\cite{planning}, deobfuscation~\cite{component-based}, invariant synthesis~\cite{inv-synth}, verified lifting~\cite{lifting}, and generating rewrite rules in SMT solvers~\cite{sygusinsmt}. 

Surprisingly, the dominant techniques for SyGuS are still based on enumerative synthesis~\cite{cvc4sy,eusolver}, where the synthesizer exhaustively enumerates a grammar, and checks resulting programs against the specification until a correct solution is found. 
Good search heuristics can lead to extreme differences in performance~\cite{cvc4sy}, so one would expect machine learning techniques to be prominent. 

In fact, in programming-by-example (PBE), where the specification is in the form of input-output (I/O) examples, the dominant heuristics are now machine-learning based~\cite{deepcoder,autopandas}. In PBE, training data is easy to generate by taking any program and generating I/O examples for it. We hypothesize that the reason these machine-learning techniques have not yet had a significant impact in SyGuS with logical specifications is the lack of easily available training data.

We present a synthesis algorithm that can learn its own heuristics via a reinforcement learning (RL) loop with relatively small quantities of training data.
We use a Monte-Carlo Tree Search (MCTS) based approach to search for solutions within the syntactic specification using policy/value functions. The policy and value functions are iteratively improved via an RL loop. To evaluate a candidate program, we use a Satisfiability Modulo Theories (SMT) solver. Implementing these techniques we were able to achieve an improvement of over $26$ percentage points ($34.7\%$ baseline vs $60.9\%$ in the best iteration) on the testing set, and similar results on the training set. Compared with the state-of-the-art tool cvc5~\cite{cvc5} we can solve $182$ problems that cvc5 cannot solve and on average $25$ more problems on the training set while solving $6$ fewer problems on the testing set.

\begin{Example}[t]
\begin{lstlisting}[language=smt,style=uclidstyle]
;; The background theory
(set-logic LIA)
;; function to be synthesized
(synth-fun f ((x Int) (y Int)) Int
    ;; syntactic constraint
    ((S Int) (B Bool)) ;; non-terminals
    ((S Int (x y 0 1 
            (+ S S)(ite B S S)))
     (B Bool ((and B B) (or B B) (not B)
              (= S S) (>= S S)))))
;; universally quantified variables
(declare-var x Int)
(declare-var y Int)
;; semantic constraints on the function
(constraint (>= (f x y) x))
(constraint (>= (f x y) y))
(constraint (or (= x (f x y))(= y (f x y))))
(check-synth)
\end{lstlisting}
    \caption{SyGuS problem, expressing the semantic constraint: $\forall x,y. f(x,y)\geq x \wedge f(x,y)\geq y \wedge (f(x,y)=x \vee f(x,y)=y)$, and a syntactic constraint.
    }
    \label{ex:sygus-ex}
\end{Example}

To obtain training data, we develop a novel method for automatically generating SyGuS problems, based on anti-unification. 
Generating training data for logical specifications is challenging. 
%
In PBE, the standard approach to generating data is to randomly generate solutions and their corresponding input-output specifications. For example, we might generate the solution $(ite (\geq x\ y)\ x\ y)$ for $f$, where $ite$ is the if-then-else operator, and the specification $f(2,1)=2$, $f(2,2)=2$, $f(1,1)=1$.  This approach cannot be easily applied to SyGuS with logical specifications because, if we randomly generate solutions, it is extremely hard to infer a meaningful logical specification that captures that solution 
without also revealing the solution in the specification. This means that any inferred specifications are unlikely to be representative of real SyGuS problems. We could also randomly generate logical specifications, but these are highly unlikely to be feasible (i.e., to have solutions). Another approach used in the literature for invariant synthesis problems is to mutate existing specifications~\cite{code2inv}, but 1) we must be careful not to mutate the specification in a way that renders the specification infeasible and 2) the approach in the literature does not generate training data with any new solutions, only new specifications. 

In contrast, the novel method we present for generating training data for SyGuS generates new, meaningful logical specifications beyond PBE. We use syntactic \emph{unification} (where assignments to variables are found to make two expressions equal) and \emph{anti-unification} (where a generalization of two terms is found) techniques on existing SMT problems to generate feasible and interesting logical specifications. This, as opposed to mutating existing specs randomly, generates feasible problems with new solutions.

The main contributions of this paper are:
\begin{squishitemize}
    \item We frame SyGuS as a tree search problem so that we can use a Monte-Carlo tree search (MCTS) based synthesis algorithm that uses machine learned policy and value predictors trained in a reinforcement learning loop.
    \item We train policy and value predictors in an RL loop.
    \item We present a method for generating SyGuS problems from first-order SMT problems using anti-unification and unification to overcome a lack of training data. 
    \item We evaluate our method on a combination of pre-existing benchmark sets from the SyGuS competition~\cite{syguscomp} and our newly generated problem set and compare it to the state-of-the-art SyGuS solver cvc5.
\end{squishitemize}


\paragraph{Related work:} 
Machine-learning based synthesis methods have been applied to PBE program synthesis problems~\cite{deepcoder,autopandas,grammarfiltering,bunel2018leveraging,DBLP:conf/iclr/OdenaSBSSD21} where the specification of correctness is a set of input-output examples, and strategy synthesis~\cite{DBLP:conf/aaai/MedeirosAL22}. 
Two approaches in literature use RL. The first uses a pre-trained policy that is then updated using deductive reasoning guided RL~\cite{dillig}. The second applies RL training an agent to interact with a Read-Eval-Print loop~\cite{repl}. Neither of these approaches are applied to synthesis with logical specifications, and so neither offers solutions to the challenges of a relatively sparse training data set.

Common algorithms for solving SyGuS are based on Oracle Guided Inductive Synthesis(OGIS)~\cite{ogis,solar2009}, which alternates between a search phase that enumerates the space of possible programs with various heuristics~\cite{cvc4sy,partial-eval,eusolver,abate2018,euphony}, and an oracle that returns feedback on candidate programs. Previously, traditional MCTS has been applied without learned guidance to expression search~\cite{DBLP:journals/ijait/Cazenave13}, and with learned guidance to program search~\cite{DBLP:journals/corr/abs-2109-00619}. The closest to our work is Euphony~\cite{euphony}, which learns a probabilistic grammar and uses this to guide an A$^*$ style search algorithm. 
Euphony is pre-trained and requires a corpus of training data problems with solutions, whereas we learn through RL and do not require solutions in our training data. 

RL~\cite{rl-inv} and graph neural networks~\cite{code2inv} have been applied to invariant synthesis, which is a version of synthesis with logical specifications, without syntactic templates. This domain also suffers from training data scarcity and data-driven approaches have to consider ways to overcome this. For instance, \cite{code2inv} take existing loop invariant specifications and mutate them in ways that are guaranteed to keep the solution the same.


%

The algorithm that we adapt to enumerative synthesis is based on AlphaZero~\cite{doi:10.1126/science.aar6404} and has successfully been applied to first-order logic theorem proving~\cite{RLtheoremproving} and combinator synthesis~\cite{DBLP:conf/lpar/Gauthier20}.

Anti-unification was first presented by Plotkin~\cite{plotkin}. It is the dual of unification, which is widely used in theorem proving, logic programming, and term re-writing~\cite{anti-unification-apps,anti-unification-logic-programming,DBLP:books/daglib/0092409}. Anti-unification has been applied to solving PBE problems~\cite{anti-unification-pbe}.

\section{Background}\label{sec:background}
\paragraph{Syntax-guided synthesis (SyGuS):} A SyGuS problem, see \prettyref{ex:sygus-ex}, is a tuple $\langle \tau,f,\syntformula,G\rangle$ where $\tau$ is  a background theory, $f$ is the function to be synthesized, $\phi$ is a quantifier-free formula and $G$ is a context-free grammar. The task is to find a body for $f$ such that $f$ is in the language specified by $G$ and the formula $\phi$ is $\tau$-valid, i.e., the formula $\exists f \forall \vec{x}. \syntformula(f)$ must be true where $\vec{x}$ is the vector of free variables in $\syntformula$. 


\paragraph{Context-free grammars:} A grammar $G$ is a tuple $G = (N, T, R, S_t)$, where $N$ is a finite set of symbols (non-terminals); $T$ is a finite set of symbols (characters of the language); $N$ and $T$ are disjoint;
$R$ is a set of production rules, where each rule is of the form $N \rightarrow (N \cup T)^*$, where $^*$ represents the Kleene star operation;
%
and $S_t$ is a symbol in $N$ (the start symbol). We use $\LanguageOfGrammar{G}$ to denote the language specified by the grammar $G$.
%



Given a context-free grammar $G = (N, T, R, S_t)$, the grammar tree is defined as follows: $S_t$ is the only root (i.e. node with no parent), and for every node $r$, $r[^1\alpha \rightarrow \beta]$ is a child of $r$ if  $\alpha \in \vars{r}$ and $\alpha \rightarrow \beta \in R$.
Note that $r[^1\alpha \rightarrow \beta]$ is the substitution of $\beta$ for $\alpha$ for the leftmost occurrence of a non-terminal variable $\alpha$ in $r$.

\begin{figure}
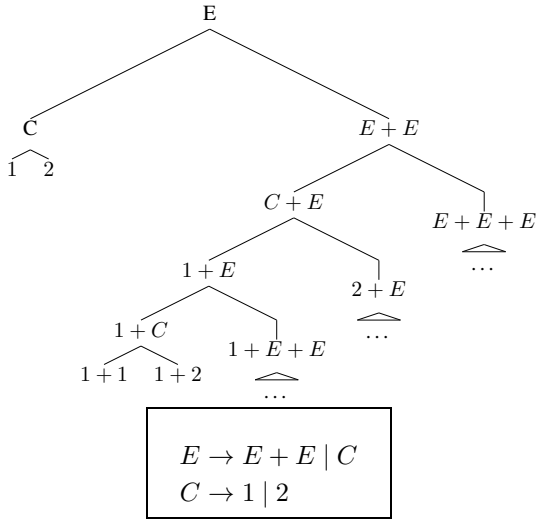

\centering
\begin{subfigure}{.4\textwidth}
\resizebox{\textwidth}{!}{
\Tree[.E    [.C    \textit{$1$} \textit{$2$}
                ]
            [.\textit{$E+E$}    [.\textit{$C + E$} 
                                    [.\textit{$1 + E$}    [.\textit{$1 + C$} \textit{$1+1$} \textit{$1+2$}
                                                          ]
                                                          [\qroof{\ldots}.\textit{$1+ E + E$}
                                                          ]
                                    ]
                                    [\qroof{\ldots}.\textit{$2 + E$}
                                                          ]
                                    ]
                    [\qroof{\ldots}.\textit{$E + E + E$}
                    ]
            ]
    ]
    }
\end{subfigure}
\begin{subfigure}{.4\textwidth}
\centering
\fbox{
\begin{minipage}{0.4\textwidth}%
\begin{align*}
E &\rightarrow E + E\ |\ C\\
C &\rightarrow 1\ |\ 2
\end{align*}
 \end{minipage}
 }
\end{subfigure}
    \caption{Grammar and corresponding grammar tree.}
    \label{fig:syntax-graph}
\end{figure}

\prettyref{fig:syntax-graph} shows an example of a \emph{grammar tree}. Every valid parse tree according to the grammar is a path from the root node to a leaf node 
of the grammar tree. Hence, every leaf node denotes a complete program and if the language of the grammar is infinite the resulting tree is also infinite. 

%
%
%
\section{Monte-Carlo Grammar Tree Search}\label{sec:MCTS}
We describe enumerative function synthesis as a single-player game. Let $P = \langle \tau, f, \phi, G \rangle$  be a SyGuS problem.
Each game state consists of a static component, the specification $\phi$ which does not change throughout the game, and a dynamic component which corresponds to the vertices in the grammar tree of $G$. 
Hence, every state $s\in S$ of the synthesis game is a pair $\langle \phi, H \rangle$ where $H$ is a partial program generated by $G$. The set of actions $A$ that can be taken in a state solely depends on the dynamic component -- the expression $H$. If $H$ is a complete program we have a final state where no further actions can be taken and we use an SMT solver to check if $H$ is a solution to $P$. If we have not reached a final state the possible actions correspond exactly to all leftmost reductions for $H$. Thus, the game tree corresponds to the grammar tree of the grammar $G$ where the root is the state $\langle \phi, \texttt{St} \rangle$ with $\texttt{St}$ being the start symbol of $G$. For all other nodes, we have that if $s = \langle \phi, H \rangle$ is a node in the game then $s' = \langle \phi, H'\rangle$ where $H'$ can be obtained from $H$ by one application of a grammar rule in $G$ is a child node of $s$

We now present the search algorithm as an agent playing the synthesis game to find solutions. In \prettyref{sec:guidance} we discuss how we incorporate machine learning with a balancing of exploration and exploitation to guide this agent.
\subsection{Algorithm}\label{sec:algorithm}
Similar to Kaliszyk et al~\cite{RLtheoremproving}, we adapt the Monte-Carlo based tree search used in AlphaZero~\cite{doi:10.1126/science.aar6404}. The algorithm consists of four main phases: \emph{Big-Steps}, \emph{Rollout}, \emph{Expansion}, and \emph{Backpropagation}. 
%
%
The algorithm's input is a SyGuS problem $P = \langle \tau, f, \phi, G \rangle$ and the output is either a solution $f \in \LanguageOfGrammar{G}$ that satisfies the constraints or a general failure (i.e. timeout). During the search we keep a record of the visit count $\visits{s}$ with the default visit count $0$ as well as other statistics that we will discuss in \prettyref{sec:guidance}. During backpropagation, these records are updated. We also keep track of whether or not a node has been expanded. In detail, the four phases work as follows:
\paragraph{Big-Steps:} As the outermost loop of the search procedure, we start at the root of the grammar tree by setting the \emph{active node} to the root of $\SyntaxGraphOfGrammar{G}$ as depicted in \prettyref{alg:bigstep}. Starting from the active node we execute the other phases of the algorithm which among other things update the visit counts of all nodes. If the rollout finds a solution, we are finished and return the solution. Otherwise, we update the active node to the most visited immediate child for the next iteration, i.e., the child node with the highest visit count.
\begin{algorithm}
\caption{\texttt{Big-Steps}}\label{alg:bigstep}
\KwData{SyGuS problem $P = \langle \tau, f, \phi, G \rangle$}
\KwResult{\texttt{fail} or solution to $P$}
$\texttt{active\_nd} \gets \TreeRoot{\SyntaxGraphOfGrammar{G}}$\;
\For{$i\gets0$ \KwTo \texttt{MAX\_BIGSTEPS}}{
  $r \gets \texttt{rollout}(\texttt{active\_nd},\; P)$\;
  \eIf{$r$ is solution}{
    \Return r\;
  } {
    \texttt{active\_nd} $\gets$ \MostVisitedChild{\texttt{active\_nd}}{\SyntaxGraphOfGrammar{G}}
  }
}
\Return \texttt{fail}
\end{algorithm}
\paragraph{Rollout, Expansion, and Backpropagation:} In this phase we perform rollouts from a given starting node in the grammar tree, i.e., traversing the grammar tree starting from that node. The purpose of this process is to collect extensive visit count data by applying  \emph{expansion} steps, \emph{checking potential solutions} where appropriate, and applying \emph{backpropagation} so that \prettyref{alg:bigstep} can make the best decision possible in choosing the next active node. As shown in \prettyref{alg:rollout1} we follow a path in the grammar tree starting from a node called \texttt{active\_nd} to the first unexpanded node by always selecting the best successor of the current node. 
In \prettyref{sec:guidance} we derive a notion of ``best'' to improve the search guidance, using the upper confidence bound on the values of the nodes. 
If the first unexpanded node is a leaf node we take the complete function it represents and the \emph{verify} function uses an SMT solver to check if the function is a solution to $P$. If it is, we return the solution, otherwise, we set the value (i.e. reward) of the node to $0$. If the node is not a leaf node, we expand the node by setting the visit count to $1$. Finally, the \emph{backpropagation} function increments the visit counts by 1 and adds the value of the current node to all nodes along the path we have taken. This whole process is repeated \texttt{MAX\_ROLLOUT} times. In the experiments, we also account for a decay factor that decreases with every big step. Since we terminate the search upon successful verification (i.e. we found a correct program) backpropagating a positive reward is not necessary.
\begin{algorithm}
\caption{\texttt{Rollout}}\label{alg:rollout1}
\KwData{\texttt{active\_nd} and $P$}
\KwResult{\texttt{fail} or solution to $P$}
\For{$i\gets0$ \KwTo \texttt{MAX\_ROLLOUTS}}{
  \texttt{current\_nd} $\gets$ \texttt{active\_nd}\;
  \texttt{sub\_path} = $[\texttt{current\_nd}]$\;
  \While{{\upshape \IsExpanded{\texttt{current\_nd}}}} {
  \tcc{select best successor node based on UCT}
    \texttt{current\_nd} $\gets$ \BestSuccessor{\texttt{current\_nd}}\;
    \tcc{append the current node to the path}
    \append{\texttt{sub\_path}}{\texttt{current\_nd}}
  }
  \eIf{\upshape \IsLeaf{\texttt{current\_nd}}}{
      \eIf{\upshape \VerifyCandidate{P}{\texttt{current\_nd}}}{
         \Return \texttt{current\_nd}\;
      } {
         \CumulativeValue{\texttt{current\_nd}} $\gets$ 0\;
         \Break\;
      }
    } {
      \Expand{\texttt{current\_nd}}{$P$}
    }
    \tcc{add the value of the current node to all nodes on the path and increment the visit counts}
    \backpropagate{\texttt{sub\_path}}{\CumulativeValue{\texttt{current\_nd}}}
}
\end{algorithm}
%
%
\subsection{Guiding the search}
\label{sec:guidance}
In this section, we introduce concepts for the \emph{quality} of \emph{nodes} and \emph{actions} as well as a method to balance the exploitation of known -- with the exploration of unknown subtrees.

Similar to AlphaZero, we use the notions of value and policy of each state and edge in the game tree of the synthesis game. 
The value function $\val{s} : S \mapsto \Reals$ maps each state in the grammar tree $s \in S$ to a real number that represents the ``quality'' of the state $s$. 
This estimates the likelihood of reaching a successful state in the subtree starting from the given state. The second heuristic that we use is the policy function $\pol{s}{a}: S \times A \mapsto \Reals$ mapping a state $s$ in the grammar tree and action $a$ to a real number, which estimates the probability of success when committing to action $a$ from state $s$. For readability we use the notation $\pol{s}{s^{\prime}}$ where $s, s^{\prime} \in S$ to denote the policy of the action $a$ that leads from state $s$ to $s^{\prime}$. In addition to the visit count as mentioned above we also keep a record of the cumulative value $\CumulativeValue{s}$ and policy 
$\pol{s}{s^{\prime}}$ for each state $s, s^{\prime} \in S$ in the search graph these values are queried and saved during the \emph{expansion} phase. During the backpropagation phase, we also add the value $\val{\mathrm{end}}$ of the last node $\mathrm{end}$ in the path to all other nodes resulting in the \emph{cumulative} value. 


\paragraph{Upper Confidence Bound for Trees:}
If we always select the subtrees with the highest average values and policies we would never explore unseen parts of the tree and gain more information about other parts of the graph. Moreover, if the value and policy are not perfect (i.e. bias against subtrees that contain the solution) we might end up repeatedly exploring a path that does not lead to a successful node. 
Conversely, if we only look for unseen paths in the search tree we do not use the information provided by the value and policy functions rendering them useless. 
Hence, our goal is to strike a balance between \emph{exploration} of unknown parts of the graph
and \emph{exploitation} of the information we have already obtained in order to reach a solution as fast as possible. 
%
To this end, we use the upper confidence bound for trees (UCT)~\cite{DBLP:conf/ecml/KocsisS06} as a heuristic to select the ``best'' child node, whilst balancing the trade-off between exploration and exploitation.
Given a parent node $p \in S$ and child node $c \in S$ the UCT value of the child node when starting in the parent node is calculated as follows:
\begin{equation*}
  \uct(p, c) = \frac{\CumulativeValue{c}}{\visits{c}} +  \gamma * \pol{p}{c} * \sqrt{\frac{\log{\visits{p}}}{\visits{c}}}\label{fun:uct}
\end{equation*}
This is the sum of two terms: an exploitation term $\frac{\CumulativeValue{c}}{\visits{c}}$ which calculates the average value per visit of the child node and an exploration term $\pol{p}{c} * \sqrt{\frac{\log{\visits{p}}}{\visits{c}}}$. The constant $\gamma$ defines the relation between exploration and exploitation (i.e. if $\gamma = 0$, we only consider exploitation).
The exploration term weighs the policy $\pol{p}{c}$ (i.e. likelihood of success when choosing the action that leads from $p$ to $c$) with a term 
that increases in value when the visit count for $p$ increases and the visit count for $c$ stagnates. Conversely, if the child $c$ visit count increases comparatively to the parent visit count the term decreases in value, and thus the exploration value decreases. In combination, this leads to child nodes with fewer visits than their siblings having a larger exploration value.
\subsection{Learning policy and value functions}\label{sec:learnedPV}

\paragraph{Features:}
To facilitate the use of machine learning models we have to design vector representations for synthesis states and actions. 
The main features we use are term walks which are parent/child pairs in the syntax tree of a term.
These syntactic features are similar to those used in first-order logic theorem provers~\cite{DBLP:conf/mkm/JakubuvU17,DBLP:conf/cade/ChvalovskyJ0U19,DBLP:journals/jar/FarberKU21,RLtheoremproving}.
We use a bag-of-words representation where the words are the hashes of all constants, variables, and term walks of size two.
This results in a feature vector of size $n$, where the entry at the $i$th position denotes the number of features with hash congruent to $i \mod n$. The constant $n$ is the hash base and describes the maximum number of distinct features.

Finally, we apply this method to the formula $\phi$ and term $H$ to obtain two feature vectors the concatenation of which is the feature vector of the state $\langle \phi, H\rangle$. Similarly, we encode the action $a$ that leads from state $s = \langle \phi, H\rangle$ to $s^{\prime} = \langle \phi, H^{\prime}\rangle$
as the concatenation of the feature vectors of $H$ and $H^{\prime}$. Thus, the feature vector for policy estimation is the concatenation of the feature vectors for $\phi$, $H$, and $H^{\prime}$.
\paragraph{Learning:}
Policy and value prediction can be described as a regression model, so many machine learning models can be used here.
However, due to the speed of tree traversal and node selection in our algorithm, the calls to the learning models are the bottlenecks, much like in theorem proving~\cite{DBLP:conf/cade/ChvalovskyJ0U19, DBLP:conf/nips/IrvingSAECU16}. Hence, choosing models with low prediction latency is vital. In our experiments, we use gradient boosted trees provided by XGBoost~\cite{Chen:2016:XST:2939672.2939785}, which have been successful in theorem proving. We discuss more details in \prettyref{sec:eval}.
\paragraph{Reinforcement Learning:}
We use an RL loop to train the policy and value models. In the first iteration, we run the synthesis procedure on all training problems using a default policy of $1$ for each action and a default value of $0.95^{\#(NT)}$ where $\#(NT)$ is the number of non-terminals in the synthesized expression. Once completed, we collect the following training data from the problems and their solutions:
\begin{description}
    \item[Policy training data:] For each pair of consecutive states $s = \langle \phi, H\rangle$ and $s^{\prime} = \langle \phi, H^{\prime}\rangle$ in each successful search path we take the pair $(\langle \phi, H, H^{\prime}\rangle, p)$ where $p$ is the number of visits of $s^\prime$ in relation to the sum of all visits of all children of $s$.
    \item [Value training data:] For each state $s = \langle \phi, H\rangle$ on every path we take the pair $(\langle \phi, H\rangle, 0.9^D)$ where $D$ is the distance from the last (i.e. successful) node as training data. In case of a failed search, we use the pair $(\langle \phi, H\rangle, 0)$.
\end{description}
At the end of every RL iteration, this data together with the data of previous iterations is used to train
new policy and value functions that will be used for the next iteration.
\section{Generating SyGuS Problems}\label{sec:data-set}
The algorithm above is trained using an RL loop that learns policy and value functions. Consequently, the larger the corpus of training data, the better the algorithm is likely to perform. Randomly generating SyGuS problems as data is not practical: if we randomly generate the specification $\phi$, it is highly unlikely that a valid $f$ exists; and if we randomly generate the solution $f$, we need to find a way to infer a meaningful specification $\phi$ that admits $f$ and does not give away the answer.
The number of benchmarks available in the linear arithmetic (LIA) category of the SyGuS competition is relatively small (less than 1000), due to the relative immaturity of the field, but there are many more first-order LIA SMT problems, so we now describe a novel technique for generating SyGuS problems from SMT problems, using syntactic unification~\cite{DBLP:journals/toplas/MartelliM82} and anti-unification~\cite{DBLP:journals/corr/abs-2302-00277,plotkin}. 
The former finds variable assignments to equations that make the terms syntactically equal. For example when applying unification to the equation $(5 \odot x) \bullet (3 \ominus c) = (y \odot 1) \bullet z$ we get the unifier $\{x \mapsto 1,y \mapsto 5, z \mapsto (3 \ominus c)\}$. If we apply the mapping of variables to the term above we get the two terms that are syntactically equal $(5 \odot 1) \bullet (3 \ominus c) = (5 \odot 1) \bullet (3 \ominus c)$.
The latter is a process that produces a generalization common to a set of symbolic expressions. For example, applying anti-unification to the two terms $(2 \bullet 1) \oplus 8$ and $(1 \bullet 3) \oplus 5$ we obtain the least general generalizer (LGG) $x \bullet  y \oplus w$. In other words, we obtain a term such that if we apply unification with either of the original terms we obtain a unifier that turns the LGG into that original term. Given a $\tau$-valid SMT problem $Q = \langle \tau, \phi \rangle$, i.e., an SMT problem that is true for all variable assignments, we generate a SyGuS problem $P$ as follows:
\begin{mylist}
    \item Heuristically select a set $S$ of sub-terms of $Q$
    \item Compute LGG $l$ of $S$ with fresh variables $x_1,\dots, x_{n}$
    \item Replace each term $t \in S$ in $Q$ with synthesis target $f(x_1,\dots, x_{n})$ using unifier of $t = l$ to determine the correct variable mapping
    \item Generate a grammar $G$ that produces all terms in $\tau$ using variables $x_1, \ldots x_{n}$. 
    \item Create SyGuS problem $P = \langle \tau, f, \phi, G \rangle$.
\end{mylist}
The LGG is one possible solution to the resulting problem $P$.
The first step has the largest impact on the quality of the resulting problem as,
if the wrong subterms are chosen, the resulting LGG might be trivial (e.g. for the terms $\{5 + 4, 9 * x\}$ the LGG is basically the identity function). In our case, we search for sub-terms that have the largest LGG as well as sub-terms of type \texttt{int}. 
Sub-terms of type \texttt{bool} often lead to trivial solutions such as \texttt{True} or \texttt{False}.
\subsection{Analysis of new training data}
We use SMT problems taken from the LIA and QFLIA (Quantifier-free Linear Integer Arithmetic) tracks of the SMT competition~\cite{smtcomp} to generate new SyGuS problems. We run cvc5~\cite{cvc5} on the new problems to group them into the following categories, in approximate order of complexity: 
    \textbf{Basic (B) problems} that cvc5 can solve and where a valid solution is a function returning a single constant or variable);
    \textbf{Straight-line (S) problems} that cvc5 can solve and are not basic but where a valid solution does not need control flow (i.e., there is a valid solution that performs simple mathematical operations);
    \textbf{control-flow (C) problems} that cvc5 can solve and where any valid solution needs control flow (i.e., a valid solution must contain if then else statements); and
    \textbf{unsolved (U) problems }that cvc5 cannot solve with a $120$s time-out.

Table~\ref{tab:data} reports the number of problems in each of these categories in our \emph{new} data-set and in the original (\emph{old}) data-set from the SyGuS competition. The SyGuS competition data comprises all benchmarks in LIA from the General and Invariant Synthesis tracks with a single synthesis function. If a benchmark has no grammar, we augment the benchmark with the same grammar as our generated data. 

After running our data generation procedure overnight, we generate 8186 new SyGuS LIA problems. We remove basic problems and filter the remaining problems to reduce the number of very similar problems, using a heuristic based on the file names and duplicate comments as these are usually indicative of the source of the SMT benchmark, and correlate with very similar problems. This gives $944$ new SyGuS LIA problems, and, in total, we have a problem set consisting of $1901$ SyGuS LIA problems. 
This more than doubles the total number of SyGuS LIA problems, and increases the number of solvable benchmarks with control flow by $10\times$. The number of problems we generate is primarily limited by the time that we run the data generation, as multiple problems can be generated from each first-order problem, and the number of publicly available SMT problems increases year-on-year year.
\begin{table}
    \centering
    \begin{tabular}{l|c|c |c|c |c }
        Data set &  \#B & \#S & \#C & \#U & Total \\
        \hline
        SyGuS-comp (old) & 290 &  93&   38 & 536 &  957  \\
        Generated  (new) & 3760  & 2495 & 1693 & 220 & 8168 \\
        Filtered new (new) & -- &  211 &  513 &  220 & 944\\ 
    \end{tabular}
    \caption{Number of problems by category: basic problems (B), straight-line problems (S), problems with control-flow (C), and unsolved problems (U).}
    \label{tab:data}
\end{table}
\section{Experimental Evaluation}\label{sec:eval}
\begin{table*}
\centering
\begin{tabular}{llllr|rr|rrrrr|rr}
    && action     &   data        &   total & \multicolumn{2}{c}{baseline}  &   \multicolumn{5}{c}{Solved in Best Iteration}                        & \multicolumn{2}{c}{cvc5} \\
 \midrule
    &&            &               &   \#    &    \#     &   \%              &        min    &     max       &  \#  mean     & \% mean   & stdev     &  \#        & \%\\
 \toprule
 \multirow{3}{*}{Exp. 1}&1  & train      &   old + new   &   1425  &   471.6   &   33.1\%         &   852         &   865         &   { 859.8} & 60.3\%  &3.97       &  834       & 58.5\% \\
 &2  & test on    &   old + new   &   476	  &   163.4   &   34.3\%         &	287           &	292           &	289           &	60.7\%   & 1.87      &  { 295} & 62.0\% \\ 
 &3  & test on    &   old         &   249   &  49.4     &	19.8\%           &	75            &	78            &	76.4          &30.7\%    &  1.14     &  { 115} & 46.2\% \\
 \midrule 
 \multirow{2}{*}{Exp. 2}&4  & train on   &   old         &   717    &  112.2   &	15.7\%           &	205           &	224           &	210.8         & 29.4\%   & 7.6       &	{ 315}  & 41.0\% \\
  &5 & test on    &   old*        &   240    &	39.8     &16.6\%	         &     69	     &  79           &	73.6         &    30.7\%&	3.58     &	{ 108} & 45.0\%  \\
 \midrule
\multirow{3}{*}{Exp. 3}&6 & train on   &   new         & 944	  &  479.6	  &50.8\%           &	874          &	886          &	{ 879.6}  &	93.2\%  &	3.37     &  727       & 77.0\% \\
  &7 & test on    &   old         & 957      &	151.8     &15.9\%           &	169	         &184	         &    175.4	     &18.3\%    &	5.97     &  { 402} & 42.0\% \\
  &8 & test on    &   old*        & 240	  &	38.6   &	16.1\%          &	45           &	50           &	47.4	     &   19.8\% &	2.51     &  { 108} & 45.0\%\\
 \bottomrule
\end{tabular}
\caption{Results summary. Training and testing sets are disjoint sets. The two old* data sets are the exact same sets.} \label{tab:summary}
\end{table*}
We implemented the Monte-Carlo tree search algorithm presented in \prettyref{sec:MCTS} with the learned policy and value using RL
as discussed in \prettyref{sec:learnedPV}. Our experimental evaluation seeks to answer the following  questions:
\begin{description}
\item[Q1:] Can MCTS with RL be used in function synthesis?
\item[Q2:] Does our data generation improve the performance?
\item[Q3:] How does MCTS compare to other techniques?
\end{description}
The implementation is written in C++ using Z3~\cite{DBLP:conf/tacas/MouraB08} as an SMT solver to check the correctness of potential candidate solutions. We use XGBoost~\cite{Chen:2016:XST:2939672.2939785} as a library to provide gradient-boosted trees for policy and value estimation. 
Running the experiments multiple times with different hyper-parameters ranging from ($10$ to $30$) we found that the best tree depths are $20$ and $25$ for value and policy. We did not experience any significant improvements by changing other hyper-parameters. In the MCTS we do $30$ big-steps and $6500$ rollouts with a decay factor of $0.98^B$ where $B$ is the number of big-steps. These were optimized for a $100$s timeout. 
As a hash base for the feature vectors, we use $2^{12} -3 = 4093$. 
Our synthesis algorithm exclusively learns from its previous iterations without external guidance or solutions. For training, we take data obtained from the previous $4$ iterations instead of all previous iterations (we tried values between $2$ and $20$). We do not train during runs on the test set.
%
%
Previously, we had a Python-based implementation where we experimented with other machine learning models such as K-Nearest Neighbours and Linear Regression as well as larger hash bases, and found that tree models outperformed other machine learning models.
%
\paragraph{Benchmarks \& Setup:} We use the SyGuS LIA data set consisting of 1901 problems described in \prettyref{sec:data-set}. We create a $75:25$ training/testing split on the data sets for each of the experiments. 
We make the data set as well as the code with Dockerfile and instructions on how to install dependencies, compile the code, and run experiments available\footnote{\url{https://zenodo.org/records/10377451}}.
The experiments were run on Amazon Elastic Compute Cloud (EC2) on a \texttt{r6a.8xlarge} instance with an AMD EPYC 7R13 CPU and 256GB of Memory running Amazon Linux 2.
We run each search procedure with a timeout of $100$ seconds. Our implementation introduces non-determinism in two ways: training of the machine learning models and a random tie break in child selection when two or more child nodes have the same scores. We ran experiments 5 times with fixed seeds.
\subsection{Results}
\begin{figure}
    \centering
    \includegraphics[scale=0.54]{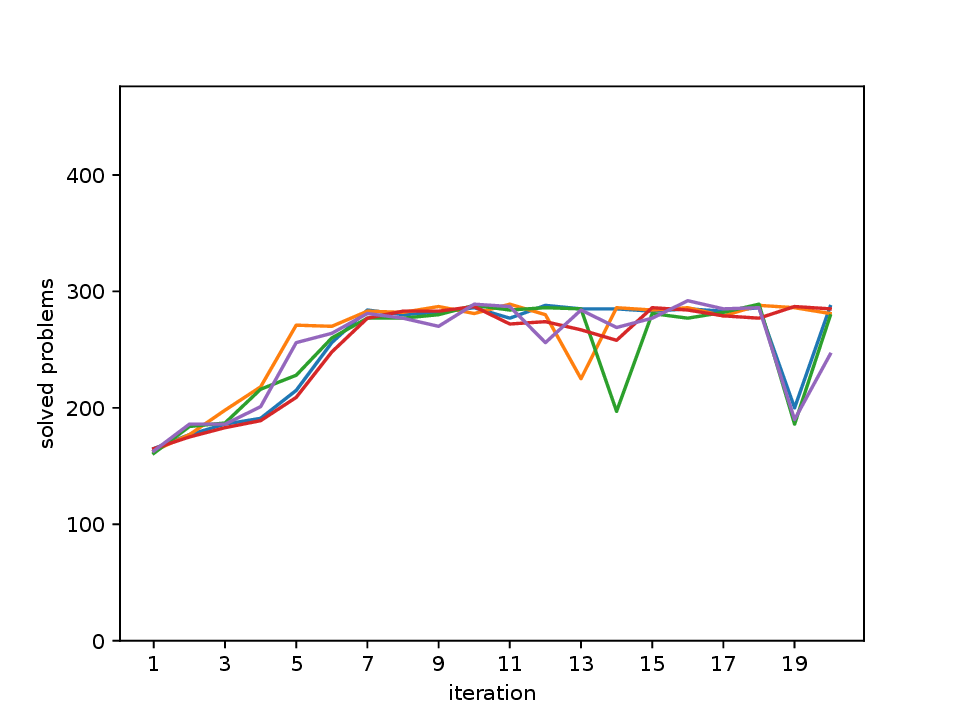}
    \caption{Number of solved problems in the testing set for each iteration in each of the $5$ experiment runs. Each line represents one experimental run.}
    \label{fig:exp-over-iterations}
\end{figure}

\paragraph{Q1: Can MCTS with RL be used in function synthesis?} We run experiments on the union of all benchmark sets resulting in $1425$ training and $476$ testing problems. The results are shown in experiment 1 (Exp. 1) of \prettyref{tab:summary}. The baseline is the first iteration of the reinforcement learning loop. In this iteration, we have not gathered any training data and have no trained guidance so use default value and policy functions. 
From the second iteration onwards, policy and value functions are trained on the data obtained from the training set in the previous iterations (no data is collected from the testing set).  
\prettyref{tab:summary} shows a comparison of the performance of the best iterations of each of the $5$ experiment runs on the training and testing set. The best iteration of each experiment on average solves $859.8$ and $289$ problems on the training and testing set respectively. This is an improvement of $27.2$ and $26.4$ percentage points compared with the first iteration for training and testing sets.
\prettyref{fig:exp-over-iterations} plots the number of solved testing problems for each iteration for each experiment.
%

\paragraph{Q2: Does our data generation improve performance?}
%
Doing a more fine-grained analysis of the experiments, we find that the testing set in the previous experiments contains $249$ problems from the old data set. Of these, between $75$ ($30.1\%$) and $78$ ($31.3\%$) are solved with a mean of $76.4$ ($30.7\%$) and a stdev of $1.14$. To compare, we conduct two additional experiments with the following data set-ups, with data sets from \emph{Old} and \emph{New} problems, as described in \prettyref{sec:data-set}, and report the results in \prettyref{tab:summary} 
%
\begin{enumerate}
    \item Train with \emph{old} and test with \emph{old} problems (Exp. 2)
    \item Train with \emph{new} and test with \emph{old} problems (Exp. 3).
\end{enumerate}
%
In Exp. 2
the trained search agent solves almost $1.85$ times as many problems as the baseline ($16.6\%$ baseline and $30.7\%$ on average in the best iteration on the testing set). 
In contrast, Exp. 3 shows that training exclusively on  new problems and testing exclusively on  old problems only improves  performance on the testing set  (i.e. old problems) by $2.4\%$ 
($15.9\%$ on baseline and $18.3\%$ on average in the best iteration). 
In line 8 we highlight the performance of the setup in Experiment 3 on the same test set as in Experiment 2, which has a similar marginal performance improvement. 
We find that the new training data does not improve the performance on the old test set compared with simply training on the old data, but as can be seen in Exp. 1 (line 3) of \prettyref{tab:summary}, it also does not diminish the performance on the old problems, whilst still enabling us to solve a significant number of the new problems that cvc5 is unable to solve. 
We hypothesize this is because our new benchmark set is not typical of the existing SyGuS benchmarks (this is also demonstrated by cvc5's failure to solve many of these examples). 
%
%
%
\paragraph{Q3: How does MCTS compare to other techniques?}
We compare our approach to cvc5~\cite{cvc5}, the SyGuS solver which performed best in the most recent SyGuS competition~\cite{syguscomp}. 
In Exp. 1 the best iteration of our tool on average solves $25.8$ problems more (min. $18$, max. $31$ more) than cvc5 in the training set and on average $6$ fewer (min. $3$, max. $8$ fewer) on the test set.
On average, we solve $41$ test set and $141$ training set problems on which cvc5 fails.
Exp. 2 and 3 show that cvc5 performs poorly compared with our synthesis tool on the newly generated benchmarks. 
In particular, when used as training set, we solve $93.2\%$ of the new problems while cvc5 only solves $77\%$. In contrast, the best iteration of our tool lags behind cvc5 by around 12 to 15 percentage points on the old data set, as shown by Exp. 2.
%
cvc5 is a mature software consisting of over 400k lines of code incorporating many decision procedures and domain-specific heuristics to improve function search. In comparison, our implementation exclusively uses the learned search heuristics presented in Sections~\ref{sec:MCTS} and \ref{sec:learnedPV}, and is able to perform comparably to cvc5 on the testing set and even outperform cvc5 in some setups.
\paragraph{Limitations}
We do not compare our approach to specialized invariant synthesis tools, which may outperform our technique on invariant benchmarks but cannot handle the general SyGuS problems. We are also unable to compare to Euphony~\cite{euphony} as the available code does not compile, and they do not report results on LIA.  Our conclusions are based on results on the SyGuS datasets, using hyperparameters tuned on these datasets, and these results may not hold in other domains. 
We also require a base set of problems to be solved in the first iteration to generate data to train from in succeeding iterations. 
Finally, our data-generation approach is only applicable to domains where 
first-order (verification) problems are available, and we wish to solve second-order (synthesis) problems.
However, this is the case in many areas such as software verification/synthesis~\cite{DBLP:journals/toplas/DavidKKL18}, unit test generation, and reactive motion planning.



%
%
%
%
%
%
\section{Conclusions and Future Work}\label{sec:futurework}
We presented a synthesis algorithm for SyGuS based on MCTS, incorporating machine-learned
policy and value predictors and guidance based on UCT, and developed a method to generate SyGuS problems from preexisting SMT problems
using anti-unification and unification.
Our approach, using simple syntactic features and gradient-boosted tree models, improves on the baseline synthesizer by over $26\%$ on training and testing sets in the best iteration, and solves benchmarks that are out of reach of state-of-the-art synthesis tools.
%
In future work, we will exploit our new data to investigate the application of other data-dependent machine learning methods to the SyGuS domain, for instance, algorithm selection and run-time prediction~\cite{DBLP:journals/corr/HealyMP17}, and we hope that others will do the same.
\paragraph{Acknowledgements:}
This work was in part supported by the
Oxford-DeepMind Graduate Scholarship, the Engineering
and Physical Sciences Research Council (EPSRC), an Amazon Research Award, and
the Austrian Science Fund (FWF) project AUTOSARD (36623).


\bibliography{sample}

\end{document}

%% file: macros.tex
\usepackage{xspace}
\usepackage{subcaption}
\usepackage{tikz}
\usepackage{todonotes}
\usepackage{listings}
\usepackage{multirow}
\usepackage{framed}
\usepackage{subcaption}
\usepackage{microtype}
\usepackage{amsfonts}

\usepackage{enumitem}
\usepackage{prettyref}
\usepackage{amsmath}
\usepackage{qtree}
\usepackage{float}

\usepackage{comment}

\usepackage{url}

\SetCommentSty{mycommfont}

\usetikzlibrary{positioning, automata, shapes.arrows, calc, shapes, arrows, trees}
\usepackage{amsmath}

\newrefformat{tab}{Table~\ref{#1}}
\newrefformat{alg}{Algorithm~\ref{#1}}
\newrefformat{fig}{Figure~\ref{#1}}
\newrefformat{sec}{Section~\ref{#1}}
\newrefformat{def}{Definition~\ref{#1}}
\newrefformat{ex}{Example~\ref{#1}}
\newrefformat{fun}{Function~\ref{#1}}
\newrefformat{eq}{Equation~\ref{#1}}
\newrefformat{term}{Term~\ref{#1}}

\RestyleAlgo{ruled}
\SetKw{Continue}{continue}
\SetKw{Break}{break}

\definecolor{codegreen}{rgb}{0,0.6,0}
\definecolor{codegray}{rgb}{0.5,0.5,0.5}
\definecolor{codepurple}{rgb}{0.58,0,0.82}
\definecolor{backcolour}{rgb}{0.95,0.95,0.92}
\definecolor{darkmagenta}{rgb}{0.55,0,0.55}

\newcommand{\syntformula}{\phi}

\newcommand{\TreeRoot}[1]{\ensuremath{\texttt{root}(#1)}}

\newcommand{\MostVisitedChild}[2]{\ensuremath{\texttt{most\_visited\_child}(#1,\, #2)}}

\newcommand{\backpropagate}[2]{\ensuremath{\texttt{backpropagate}(#1,\, #2)}}

\newcommand{\SyntaxGraphOfGrammar}[1]{\ensuremath{GT(#1)}}

\newcommand{\VisitsSymbol}{\ensuremath{\text{N}}}
\newcommand{\visits}[1]{\ensuremath{\VisitsSymbol(#1)}}

\newcommand{\IsExpanded}[1]{\ensuremath{\texttt{is\_expanded}(#1)}}

\newcommand{\BestSuccessor}[1]{\ensuremath{\texttt{best\_successor}(#1)}}

\newcommand{\IsLeaf}[1]{\ensuremath{\texttt{isLeaf}(#1)}}
\newcommand{\CumulativeValueSymbol}{\ensuremath{\text{W}}}
\newcommand{\CumulativeValue}[1]{\ensuremath{\CumulativeValueSymbol(#1)}}

\newcommand{\Expand}[2]{\ensuremath{\texttt{expand}(#1,\, #2)}}

\newcommand{\LanguageOfGrammar}[1]{\ensuremath{L^#1}}

\newcommand{\vars}[1]{\ensuremath{\text{vars}(#1)}}

\newcommand{\val}[1]{\ensuremath{\nu(#1)}}
\newcommand{\pol}[2]{\ensuremath{\pi(#1,#2)}}

\newcommand{\append}[2]{\ensuremath{\texttt{append}(#1,#2)}}

\newcommand{\VerifyCandidate}[2]{\ensuremath{\texttt{verify}(#1,#2)}}

\DeclareMathOperator*{\uct}{uct}

\newcommand{\Reals}{\mathbb{R}}

\lstdefinelanguage{uclid}{
  sensitive = true,
  keywords={module, forall, exists, Lambda, if, else, assert, assume, havoc,
            for, range, skip, case, esac, init, next, control, function, procedure,
            returns, call, define, type, var, input, output, const, property,
            invariant, synthesis, grammar, requires, ensures, modifies, instance, axiom, 
            enum, record, integer, boolean, true, false},
  numbers=left,
  numberstyle=\footnotesize,
  stepnumber=1,
  numbersep=8pt,
  showstringspaces=false,
  breaklines=true,
  frame=top,
  comment=[l]{//},
  morecomment=[s]{/*}{*/},
}

\lstdefinestyle{uclidstyle}{
  backgroundcolor=\color{backcolour},
  commentstyle=\color{codegreen},
  keywordstyle=\color{magenta},
  numberstyle=\tiny\color{codegray},
  stringstyle=\color{codepurple},
  basicstyle=\footnotesize,
  breakatwhitespace=false,
  basicstyle=\footnotesize\ttfamily,
  breaklines=true,
  captionpos=b,
  keepspaces=true,
  numbers=left,
  numbersep=5pt,
  showspaces=false,
  showstringspaces=false,
  showtabs=false,
  tabsize=2,
  frame=shadowbox
}

\lstdefinelanguage{alg}{
  sensitive = true,
  keywords={algorithm, input, output, return, if, else, for, while},
  numberstyle=\footnotesize,
  numbersep=9pt,
  showstringspaces=false,
  breaklines=true,
  comment=[l]{//},
  morecomment=[s]{/*}{*/}
}

\lstdefinestyle{grammar}{
  belowcaptionskip=1\baselineskip,
  basicstyle=\rmfamily\mdseries\footnotesize,
  breaklines=true,
  language=alg,
  xleftmargin=\parindent,
  showstringspaces=false,
  mathescape=true,
  numberstyle=\tiny,
  keywordstyle=\bfseries
}

\definecolor{codegreen}{RGB}{51,135,97}
\definecolor{codepurple}{RGB}{93, 49, 122}
\definecolor{codered}{RGB}{128, 18, 45}
\definecolor{backcolour}{rgb}{1.0,1.0,1.0}

\lstdefinelanguage{smt}{
  sensitive = false,
  alsoletter=-,
  keywords=[1]{declare-var,set-logic,synth-fun,check-synth, fun, rec, assert, sat, blocking, define, constraint, oracle},
  keywordstyle=[1]\color{codepurple},
  keywords=[2]{div, mod, ite, and},
  keywordstyle=[2]\color{codegreen},
  keywords=[3]{Int, Bool},
  keywordstyle=[3]\color{codered},
  numbers=none,
  stepnumber=1,
  numbersep=8pt,
  showstringspaces=false,
  breaklines=true,
  comment=[l]{;},
  basicstyle=\footnotesize
}

\lstdefinelanguage{smtmini}{
  sensitive = false,
  keywords=[1]{declare, fun, synth, rec, check, assert, sat, blocking, define, constraint, oracle},
  keywordstyle=[1]\color{codepurple},
  keywords=[2]{div, mod, ite, and},
  keywordstyle=[2]\color{codegreen},
  keywords=[3]{Int, Bool},
  keywordstyle=[3]\color{codered},
  numbers=none,
  stepnumber=1,
  numbersep=8pt,
  showstringspaces=false,
  breaklines=true,
  comment=[l]{;},
  basicstyle=\tiny
}

\newcounter{myctr}
\newenvironment{mylist}{\begin{list}{\arabic{myctr}.}
{\usecounter{myctr}
\setlength{\topsep}{1mm}\setlength{\itemsep}{0.25mm}
\setlength{\parsep}{0.1mm}
\setlength{\itemindent}{2mm}\setlength{\partopsep}{0mm}
\setlength{\labelwidth}{15mm}
\setlength{\leftmargin}{4mm}}}{\end{list}}

\newenvironment{squishitemize}{\begin{list}{$\bullet$}
{\setlength{\topsep}{1mm}\setlength{\itemsep}{0.25mm}
\setlength{\parsep}{0.1mm}
\setlength{\itemindent}{0mm}\setlength{\partopsep}{0mm}
\setlength{\labelwidth}{15mm}
\setlength{\leftmargin}{4mm}}}{\end{list}}

\newfloat{Example}{hbtp}{lop}